%% file: main.tex
\documentclass[nohyperref]{article}

\input{math_commands.tex}

\usepackage{hyperref}
\usepackage{url}
\usepackage{booktabs, multirow}
\usepackage{soul}
\usepackage{changepage,threeparttable}
\usepackage{makecell}
\usepackage{graphicx}
\usepackage{xspace}
\usepackage{tabularx,colortbl}
\usepackage{adjustbox}

\usepackage{enumitem}  %

\usepackage{xcolor,colortbl}

\usepackage[accepted]{icml2023}
\usepackage{amsmath}
\usepackage{amssymb}
\usepackage{mathtools}
\usepackage{amsthm}

\usepackage[capitalize,noabbrev]{cleveref}

\theoremstyle{plain}

\theoremstyle{definition}

\theoremstyle{remark}

\definecolor{crimsonglory}{rgb}{0.75, 0.0, 0.2}

\newcolumntype{d}{>{\columncolor{crimsonglory!15}}c}

\newcommand{\p}[1]{\vspace{1mm}\noindent\textbf{#1}}

\newcommand{\ie}{\textit{i}.\textit{e}.}
\newcommand{\eg}{\textit{e}.\textit{g}.}

\icmltitlerunning{Towards Better Few-Shot and Finetuning Performance with Forgetful Causal Language Models}
\newcommand{\ours}{{FCM}\xspace}
\newcommand{\oursext}{{T-FCM}\xspace}

\begin{document}

\twocolumn[
\icmltitle{Towards Better Few-Shot and Finetuning Performance \\ with Forgetful Causal Language Models}

\icmlsetsymbol{equal}{*}

\begin{icmlauthorlist}
\icmlauthor{Hao Liu}{equal,b,g}
\icmlauthor{Xinyang Geng}{equal,b,g}
\icmlauthor{Lisa Lee}{g}
\icmlauthor{Igor Mordatch}{g} \\
\icmlauthor{Sergey Levine}{b}
\icmlauthor{Sharan Narang}{g}
\icmlauthor{Pieter Abbeel}{b}
\end{icmlauthorlist}

\icmlaffiliation{b}{UC Berkeley}
\icmlaffiliation{g}{Google Research, Brain Team}

\icmlcorrespondingauthor{Hao Liu}{hao.liu@berkeley.edu}
\icmlcorrespondingauthor{Xinyang Geng}{young.geng@berkeley.edu}

\icmlkeywords{Machine Learning, ICML}

\vskip 0.3in
]

\printAffiliationsAndNotice{\icmlEqualContribution} %

\begin{abstract}
Large language models (LLM) trained using the next-token-prediction objective, such as GPT3 and PaLM, have revolutionized natural language processing in recent years by showing impressive zero-shot and few-shot capabilities across a wide range of tasks. In this work, we propose a simple technique that significantly boosts the performance of LLMs without adding computational cost. Our key observation is that, by performing the next token prediction task with randomly selected past tokens masked out, we can improve the quality of the learned representations for downstream language understanding tasks. We hypothesize that randomly masking past tokens prevents over-attending to recent tokens and encourages attention to tokens in the distant past. We find that our method, Forgetful Causal Masking (FCM), significantly improves both few-shot and finetuning performance of PaLM. We further consider a simple extension, T-FCM, which introduces bidirectional context to causal language model without altering the sequence order, and further improves finetuning performance.

\end{abstract}

\section{Introduction}
Language model (LM) pre-training has substantially advanced the state-of-the-art across a variety of natural language processing tasks~\citep{Peters2018DeepCW, devlin2018bert, brown2020language, chowdhery2022palm} and related fields including image generation, reasoning, and  code generation~\citep{alayrac2022flamingo, lewkowycz2022solving, saharia2022photorealistic, chen2021evaluating}.
Prior work on pre-training have focused on mixing different choices of architecture (\eg, encoder-only, decoder-only, or encoder-decoder)
with different objective functions (\eg, masking or causal language modeling).
For example, masked encoder-only models such as BERT~\citep{devlin2018bert} and RoBERTa~\citep{liu2019roberta} excel in discriminative finetuning tasks such as classification. Similarly, masked encoder-decoder models such as BART~\citep{lewis2019bart} and T5~\citep{roberts2019exploring} perform well on both discriminative and generative finetuning.
While masked language modeling is effective for finetuning and removes the need for task-specific architectures, %
its major limitation %
is that there is still a need for task-specific datasets and task-specific finetuning.
On the other hand, decoder-only causal language models remove such limitations. In fact, they are capable of zero-shot and few-shot adaptation without the need for finetuning, by simply prompting the model with appropriate strings to control the generated outputs, as shown in GPT3~\citep{brown2020language} and PaLM~\citep{chowdhery2022palm}.

Driven by its impressive zero-shot and few-shot abilities, 
there has been more work on scaling causal decoder-only architectures
~\citep{zhang2022opt, black2022gptneox20b, brown2020language, chowdhery2022palm} compared to encoder-based architectures,
and there has been significant interests in studying such models in various contexts~\citep{hoffmann2022training, wei2022chain, li2021prefix, ahn2022say, chen2021evaluating}.
However, such decoder-only models are still limited by their imperfect zero-shot and few-shot adaptation compared to human performance, and their %
relatively inferior finetuning performance compared to masked language modeling.

\begin{figure*}[!t]
    \centering
    \setlength{\tabcolsep}{0pt}
    \includegraphics[width=.99\textwidth]{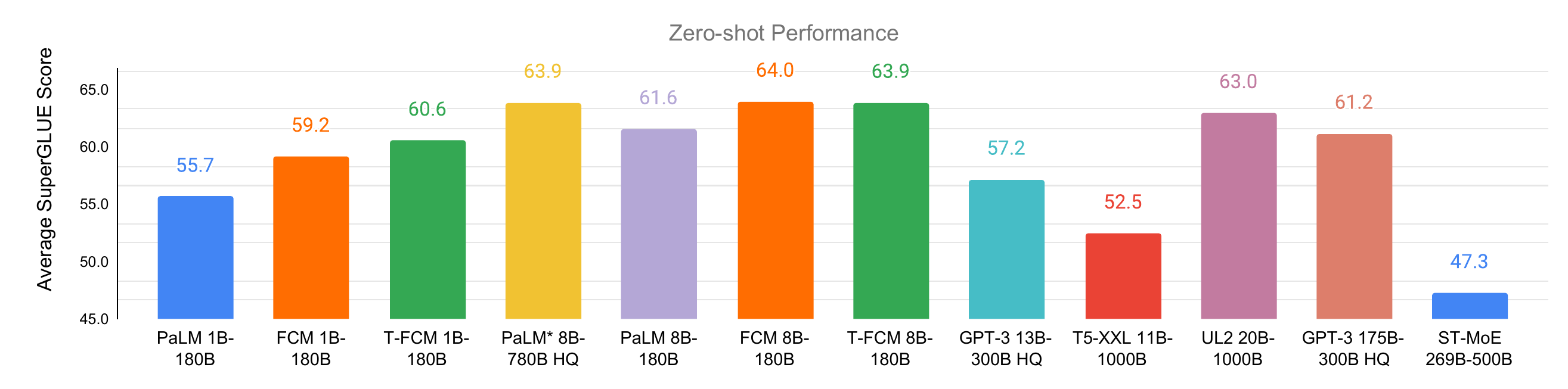}\\
    \begin{tabular}[t]{cc}
     \includegraphics[width=.49\textwidth]{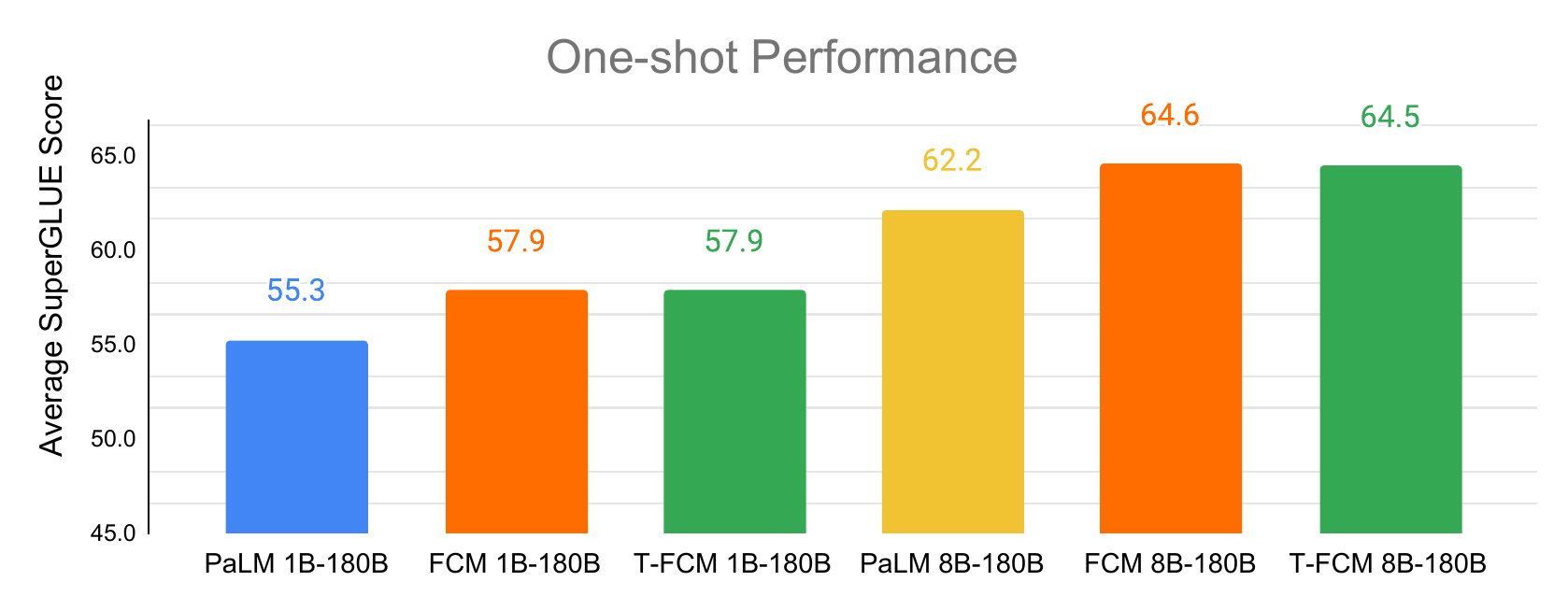}  & \includegraphics[width=.49\textwidth]{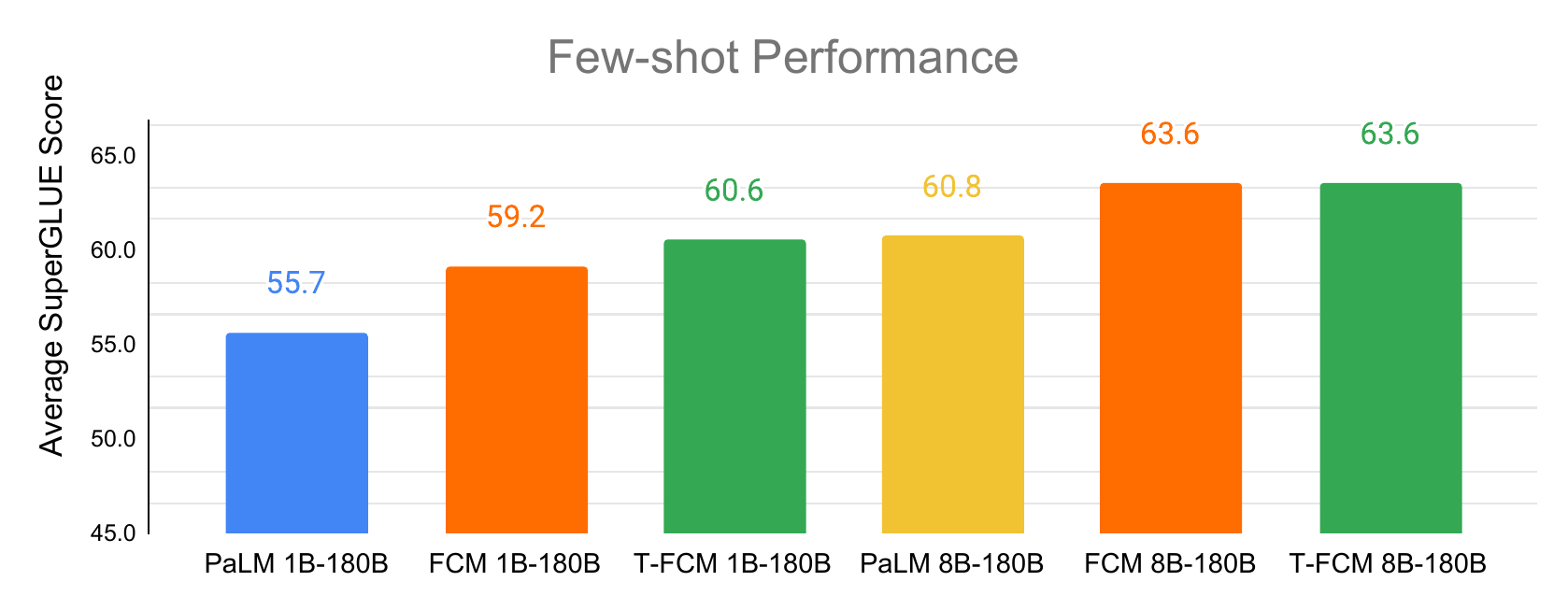} \\
     \includegraphics[width=.49\textwidth]{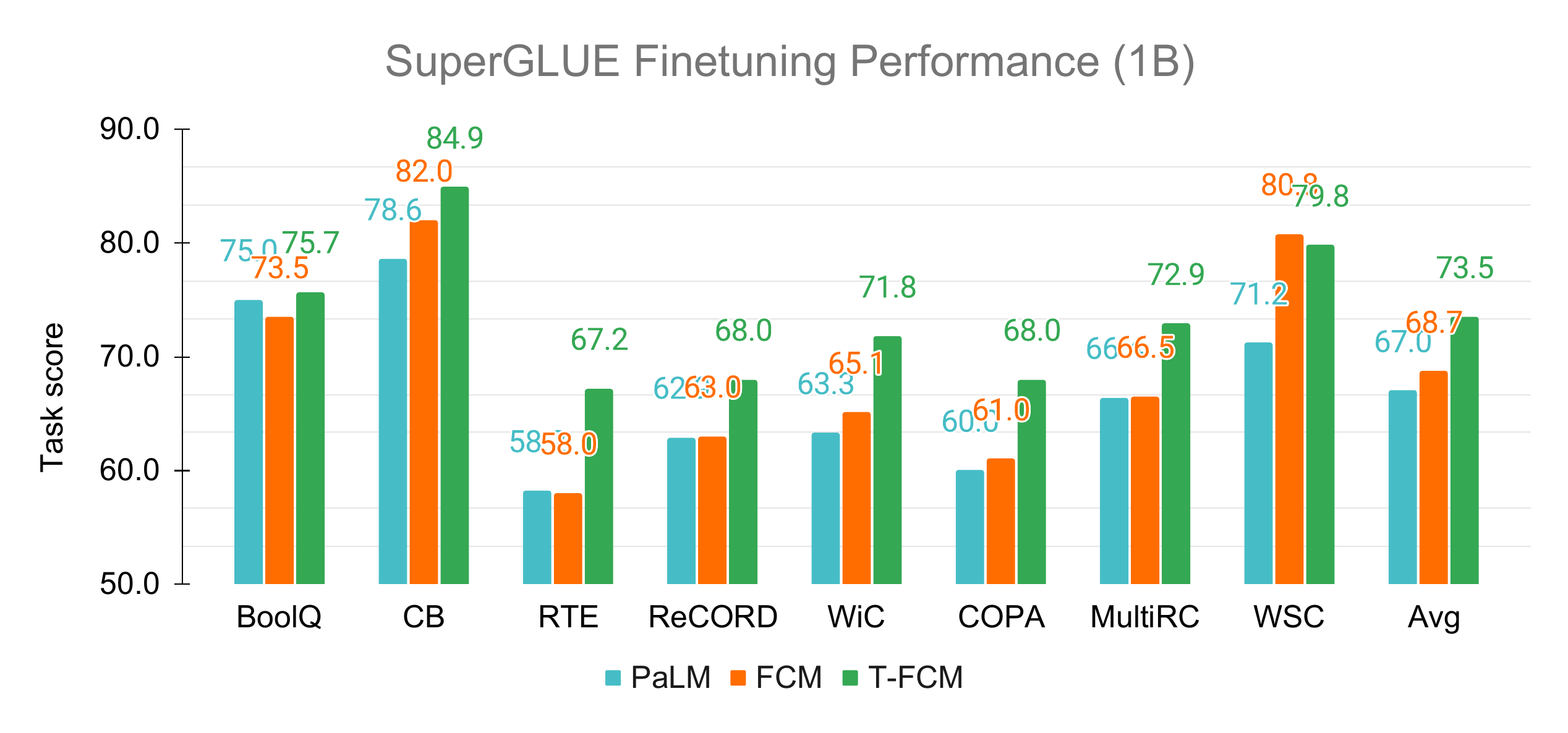} & \includegraphics[width=.49\textwidth]{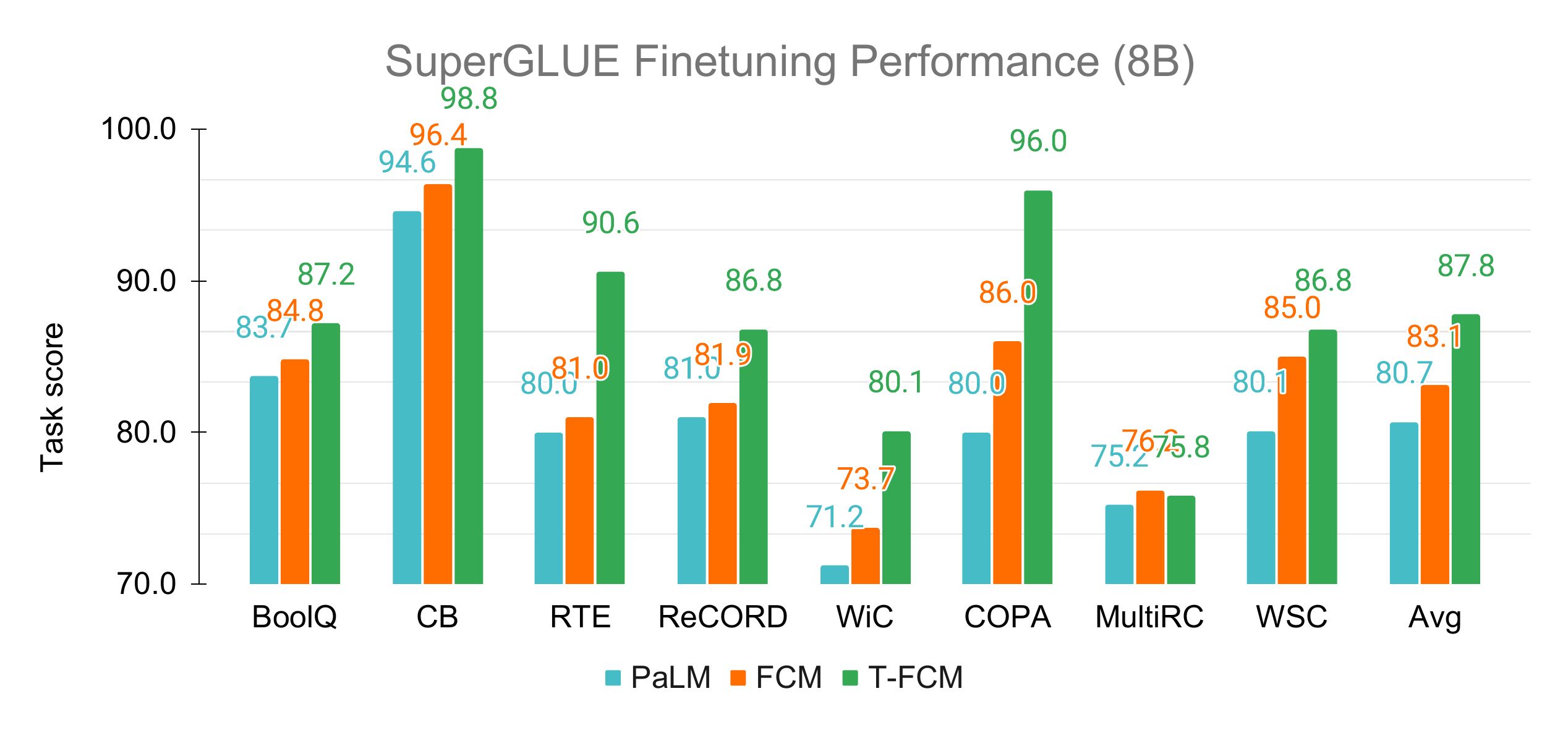} \\
    \end{tabular}
    \caption{\ours and \oursext outperform PaLM in zero- and few-shot as well as finetuning tasks. We report the averaged scores in each category. Scores are averaged over 3 evaluation random seeds. 
    \textbf{Top}:\; We compare zero-shot, one-shot, and five-shot average performance on the SuperGLUE benchmark of different model sizes and dataset sizes. PaLM$^\star$ 8B-780B HQ denotes the published results of 8B model trained on 780B tokens from high quality datasets, PaLM 8B-180B denotes the same setup but trained on 180B tokens from C4 dataset, and \ours 8B-180B denote the same 8B model trained on 180B tokens from C4 dataset using \ours as objective. 
    \textbf{Bottom}:\; We compare finetuning performance on SuperGLUE for 1B model size (left) and 8B model size (right). %
    \oursext, a simple extension of \ours, further boosts finetuning performance significantly while achieving similar few-shot performance as \ours.
    }
    \label{fig:group_highlight}
\end{figure*}

To address the above challenges, prior work have proposed to combine masked modeling with causal language modeling~\citep{dong2019unified, wang2022language, tay2022unifying, du2022glm} to bring the benefit of masked modeling to causal language models while retaining their zero-shot ability. 
However, such approaches typically introduce extra computation and parameters or require using a sophisticated attention masking strategy which hinders practical usages~\citep{yang2019xlnet, tay2022unifying}. 
Moreover, they typically train encoder-decoder models which are not naturally suitable for zero- and few-shot inference tasks compared with decoder-only causal language models and are still outperformed by causal language models~\citep{sanh2022multitask, brown2020language, chowdhery2022palm}.
In order to further improve causal language models few-shot abilities, some works proposed better prompt engineering methods~\citep{liu2021pretrain, lester2021power, ling2017program, wei2022chain, li2021prefix} or better finetuning methods~\citep{mishra2022cross, wei2022finetuned, sanh2022multitask}.
Prompt-based methods are sensitive to design~\citep{lester2021power, liu2021pretrain}, while finetuning-based approaches typically require a huge amount of supervision to work with as shown in~\citet{sanh2022multitask}.
In addition, such methods can only improve pre-trained model and are unable to improve pre-training.

In this work, we propose a pre-training approach that does not incur any extra computation cost or parameters, to improve few-shot and zero-shot performance, as well as representation learning of causal language models.
Our key observation is that, by performing next token prediction task with randomly selected past tokens masked out, we can improve the quality of the learned representations for downstream language understanding tasks. 
Our method, Forgetful Causal Masking (\ours), can be efficiently implemented by randomly masking input tokens in the causal language model.
Applying our method to PaLM~\citep{chowdhery2022palm}, a state-of-the-art causal language model, we see significant improvement on the SuperGLUE~\citep{sarlin2020superglue} benchmark: our method significantly improves the 1B-model-size PaLM's zero-shot performance from 55.7 to 59.2 and improves the 8B-model-size PaLM's zero-shot performance from 61.6 to 64.0.
We further evaluate \ours on a diverse suite of NLP tasks from~\citet{brown2020language}, and observe improvements in few-shot learning on most tasks.
In addition, \ours improves representation learning, as shown in our SuperGLUE finetuning experimental results, where our method improves 1B parameter PaLM model's finetuneing performance from 67.0 to 68.7, and our method improves 8B parameters PaLM model's finetuning performance on all 8 SuperGLUE tasks, improving the score from 80.7 to 83.1. 
We also propose an extension of our method called Two-Pass \ours (\oursext), which applies \ours %
twice on a replicated input sequence. In doing so, \oursext effectively makes causal language model see bidirectional context without altering sequence ordering. While this adds extra computation cost during training, %
\oursext further boosts finetuning performance without hurting few-shot results, improving the score from 80.7 to to 87.8 (8B) and 67.0 to 73.5 (1B).

\p{Contributions.} We highlight the contributions of our paper below: 
\begin{itemize}
    \item We present \ours, a simple and scalable pre-training methodology for causal language modeling. We provide the empirical evaluation of \ours on a suite of few-shot and finetuning benchmarks.
    \item We show that \ours is highly effective at improving zero-shot and few-shot learning results, outperforms strong baselines including PaLM and UL2, improving the average SuperGLUE score of 8 billion parameters PaLM from 61.6 to 64.0, and improving PaLM on a wide range of 19 NLP tasks.
    \item In addition to few-shot learning, we demonstrate that \ours significantly helps with finetuning to downstream tasks, improving the performance of 8 billion parameters PaLM on 8 out of 8 SuperGLUE tasks and the average SuperGLUE score from 80.7 to 83.1.
    \item We propose Two-Pass \ours (\oursext), a simple yet effective extension of \ours that introduces bidirectional context to causal language models without altering the sequence order. We observe that \oursext further boosts finetuning performance on SuperGLUE score from 80.7 to 87.8 without affecting few-shot learning performance.
\end{itemize}

\section{Method}

\begin{figure*}[!htbp]
    \centering
    \includegraphics[width=.99\textwidth]{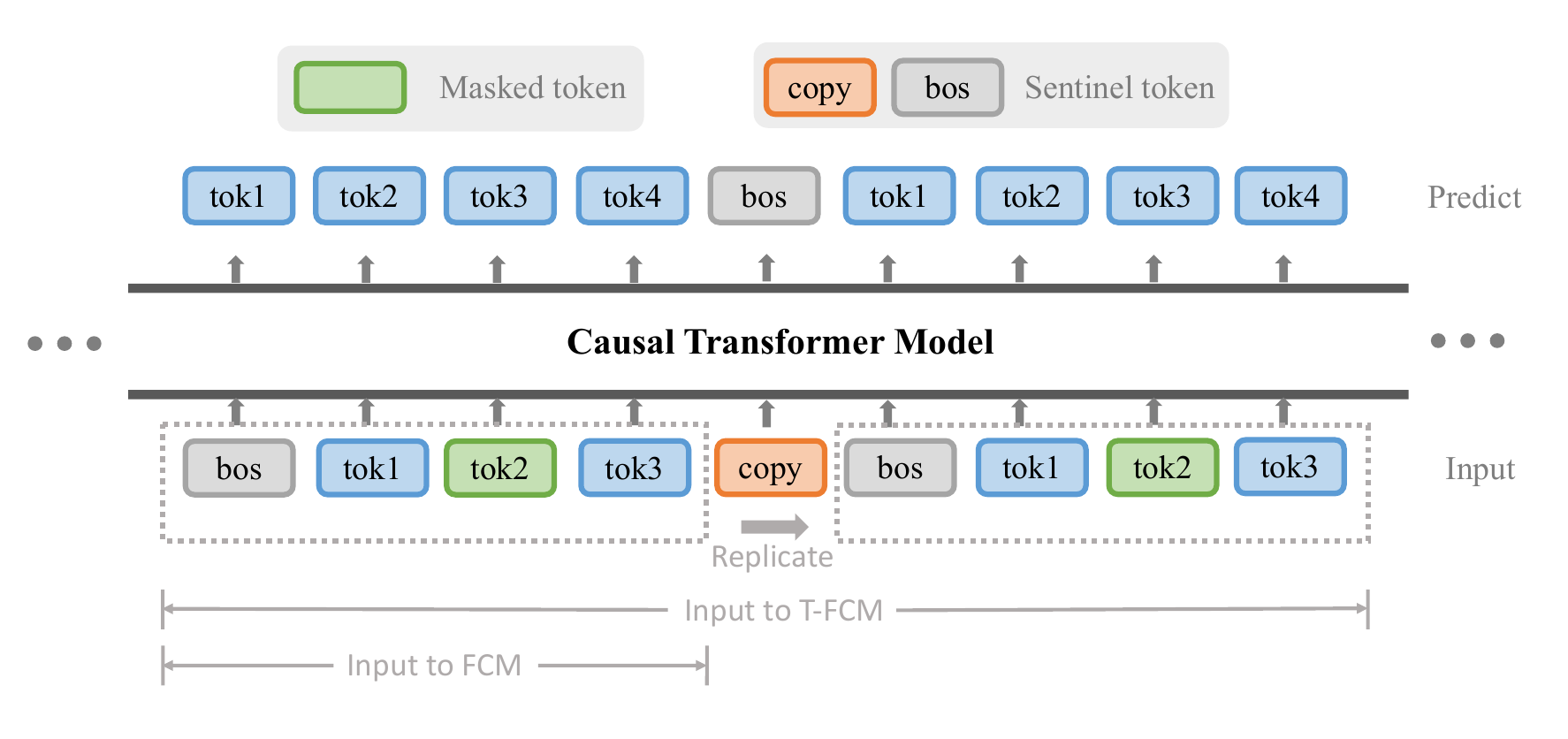}
    \vspace{-0.5em}
    \caption{Illustration of \ours and its extension \oursext. Given a causal language model, each token's prediction is conditioned on previous embeddings that are not masked. 
    Sampled tokens (green) are masked using an attention mask in each self-attention layer. %
    The loss is applied to and averaged over all tokens in the sequence. 
    In the example above, \ours predicts tok3 conditioned on [bos, tok1], and predicts tok4 conditioned on [bos, tok1, tok3]. \oursext inputs the entire replicated sequence, and predicts tok2 conditioned on [bos, tok1, tok3] on the left and [copy, bos, tok1] on the right, effectively achieving bidirectional language modeling in causal language models,  without altering the order of the input sequence tokens.
    }
    \label{fig:model}
\end{figure*}

\subsection{Pre-training Objective}\label{sec:objective}

\p{Forgetful Causal Masking (\ours).}
\ours uses a standard causal, decoder-only Transformer model architecture~\citep{ashish2017attention}, \ie, each timestep can only attend to itself and past timesteps. We illustrate \ours in Figure~\ref{fig:model}.
Given an input text $\mathbf{x}=[x_1,\cdots,x_n]$, the standard causal language modeling objective is defined to maximize the log likelihood of $\boldsymbol{x}$ autoregressively:
\begin{align}
    \log p(\mathbf{x}) 
    & = \log \prod\limits_{i=1}^{n}p(x_i \vert x_1, x_2, \ldots, x_{i-1}) \nonumber \\ 
    & = \log \prod\limits_{i=1}^{n} p(x_i \vert \mathbf{x}_{< i } )
    := \log \prod\limits_{i=1}^{n} p(x_i \vert [x_j]_{j=0}^{i-1} ).
\end{align}

In \ours, we randomly sample a mask ratio from $m \sim [0, \eta]$ where $\eta \in [0, 1]$ is a fixed maximum mask ratio. We use $\eta = 0.15$ throughout the experiments unless otherwise mentioned. %
The model is asked to predict each token $x_i \in \mathbf{x}$ , and can only attend to tokens in $\mathbf{x}_{< i }$ that are not sampled. 
Concretely, the \ours objective is given by:
\begin{align}
    \log p(\mathbf{x}) = \log \prod\limits_{i=1}^{n} p(x_i \vert [I[m_j > \eta] \cdot x_j]_{j=0}^{i-1} ),
\end{align}
where $m_j \sim \mathcal{U}(0, 1)$.
This can be efficiently implemented by combining it with causal attention mask. 
While applying random masking to the token sequence, we always exclude the special \texttt{BOS} (`beginning of sentence') token at the beginning of each sequence, so that %
the model is aware of the beginning of a sentence. Moreover, keeping the \texttt{BOS} token unmasked helps with training stability because it ensures that there is at least one token unmasked without changing the semantic meaning of the sequence. For example, when predicting token $x_t$ for small $t$, it is possible that all tokens $[x_1, ..., x_{t-1}]$ are masked, which can cause instability in the training loss. We found that this technique enables us to train arbitrary high mask ratios without incurring instability. 

\p{Two-Pass FCM (\oursext).} Prior work has discovered that masked language models have better finetuning performance than similar size or bigger causal language models~\citep[see, \eg,][inter alia]{wang2022language, tay2022unifying}. One hypothesis for this performance gap is that masked language models can use bidirectional context during training, while causal language models cannot. %
To bridge this gap, we propose Two-Pass FCM (\oursext) to introduce bidirectional context into causal language models during training. \oursext simply replicates the input sequence twice, where the first pass is identical to causal language modeling, and the second pass is similar to masked language modeling since the model can attend to masked future tokens by looking into the first pass.

During training, \oursext introduces an additional sentinel token [copy] to let the model know that a copied sequence begins.  
In practice, we found it important to not apply loss on predicting [bos] from [copy] token, otherwise it destabilizes training. The reason is that the position of [copy] is arbitrary, hence predicting [bos] from [copy] is not well-defined.

\p{Computation cost}: \ours is a simple yet effective technique that incurs no additional computation cost to train. On the other hand, \oursext introduces additional training cost, although inference cost at test time remains the same. \oursext achieves better finetuning performance than \ours and comparable few-shot performance as \ours, making it an effective variant of \ours especially for practical applications where representations are ubiquitous.
We note that while self-attention has a quadratic cost, the compute cost of large language models is not dominated by self-attention. In our experiments, we observe that \oursext is about twice slower for 1B model (25 hours $\rightarrow$ 51 hours) and 50\% slower for 8B model (100 hours $\rightarrow$ 140 hours) to train than \ours instead of four times slower.

\subsection{Model Architecture}\label{sec:archiecture}
We use the same model and architecture as PaLM~\citep{chowdhery2022palm}, including the modified activation~\citep{shazeer2020glu}, multi-query attention~\citep{shazeer2019fast}, parallel layers~\citep{gpt-j} and RoPE embeddings~\citep{su2021roformer}, %
with the exception that we use SentencePiece~\citep{kudo2018sentence} vocabulary with 32K tokens from C4~\citep{raffel2020exploring}.
To study the dependence of \ours %
on model size, we train 3 different sizes of the model, ranging over three orders of magnitude from 125 million parameters, to 1 billion parameters, and to 8 billion parameters (see Table~\ref{tab:model_size}).

\begin{table}[h!]
\caption{Architecture details of different sized models. 
We list the number of layers, $d_\textrm{model}$, the number of attention heads and attention head size, training batch size, and sequence length.
The feed-forward size $d_\textrm{ff}$ is always $4 \times d_\textrm{model}$ and attention head size is always 256.
}
\label{tab:model_size}
\begin{center}
\scriptsize
\begin{tabular}{ lccccc } 
\toprule
Model & Layers   & \# of heads   &  $d_\textrm{model}$ & Batch size & Seq len  \\ 
\midrule
1B & $16$ & $8$  & $2048$ & $1024$ & $1024$ \\
8B & $32$ & $16$  & $4096$ & $1024$ & $1024$ \\
\bottomrule
\end{tabular}
\end{center}
\end{table}

\p{Training and inference.}
Our training optimizer follows PaLM, and use the Adafactor optimizer \citep{shazeer2018adafactor} which scales the learning rate by the root-mean-square of the parameter matrix.
We use a learning rate of $0.01$ for the first 10,000 steps, which is then decayed at a rate of $1/\sqrt{k}$, where $k$ is the step number. We train with momentum of ${\beta}_{1} = 0.9$. The second-order moment interpolation value is computed as ${\beta}_{2} = 1.0 - k^{-0.8}$, where $k$ is the step number. 
Following typical large Transformer models training as in PaLM and GPT-3, models are trained without dropout, and dropout of 0.1 is used for finetuning. Our training and inference codebase is based on JAX and T5X, and all models are trained on TPU v4 Pods. 
The sequence length of \oursext is $2\times$ of conventional causal language models, although its effective context length is the same. 
At test time, \ours and \oursext are evaluated the same way as conventional causal language models.
We report results averaged over three random seeds. For results of baselines, we choose and report the best published results to compare against \ours. We use exactly the same batch size, learning rate, and training hyperparameters for PaLM and \ours. More details on hyperparameters, compute infrastructure, and training time are provided in Appendix~\ref{sec:appendix}.

\section{Main results}

\newcommand{\shot}[1]{\tiny (#1)}
\begin{table*}[!ht]
    \footnotesize
    \caption{Results obtained by the \ours and \oursext 1B and 8B model across NLP benchmarks. We use the same setup as in \citet{brown2020language, chowdhery2022palm}, including the splits for each task.
    }
    \label{tab:few_shot_all_tasks}
    \vspace{0.5em}
    \setlength{\tabcolsep}{3.5pt}
    \centering
    \scalebox{1.0}{
    \begin{tabular}{p{2cm}||ccc|ccc|ccc}
    \toprule
    & \multicolumn{3}{c}{\bf Zero-shot} & \multicolumn{3}{|c|}{\bf One-shot} &  \multicolumn{3}{c}{\bf Few-shot} \\
    \cmidrule(l{3pt}r{3pt}r{3pt}){2-4} \cmidrule(l{3pt}r{3pt}r{3pt}){5-7} \cmidrule(l{3pt}r{3pt}r{3pt}){8-10}
    \bf Task & 
    PaLM \scriptsize{1B} &
    \ours \scriptsize{1B} &
    \oursext \scriptsize{1B} &
    PaLM \scriptsize{1B} &
    \ours \scriptsize{1B} &
    \oursext \scriptsize{1B} &
    PaLM \scriptsize{1B} &
    \ours \scriptsize{1B} &
    \oursext \scriptsize{1B}
    \\
    \midrule
    Lambada (EM) &42.4 &\bf{43.5} &43.0 &48.9 &\bf{49.5} &49.3 &48.2 &\bf{49.7} &48.0 \\
    StoryCloze &\bf{68.8} &68.2 &68.1 &\bf{67.3} &66.9 &66.0 &65.9 &\bf{66.7} &66.4 \\
    PIQA &72.0 &\bf{72.1} &72.0 &71.0 &\bf{71.6} &71.0 &\bf{72.0} &71.6 &71.6 \\
    ARC-e &\bf{46.2} &45.6 &45.8 &\bf{48.0} &45.9 &45.3 &\bf{50.2} &48.2 &48.9 \\
    ARC-c &25.8 &\bf{27.7} &26.9 &26.3 &27.2 &\bf{27.9} &26.5 &\bf{28.1} &28.0 \\
    OpenbookQA &45.8 &\bf{46.4} &45.6 &\bf{45.0} &43.2 &44.0 &42.6 &\bf{43.6} &42.8 \\
    Winograd &67.0 &\bf{70.0} &68.9 &67.0 &\bf{67.4} &67.2 &64.8 &\bf{70.0} &65.9 \\
    Winogrande &54.0 &\bf{54.5} &54.1 &54.0 &\bf{55.8} &55.6 &53.6 &\bf{55.0} &54.9 \\
    BoolQ &45.9 &\bf{56.0} &54.3 &48.3 &\bf{52.6} &51.6 &48.1 &46.8 &\bf{52.8} \\
    Copa &72.0 &\bf{74.0} &74.0 &72.0 &\bf{73.0} &68.0 &70.0 &\bf{72.0} &71.0 \\
    RTE &50.9 &53.8 &\bf{58.2} &53.1 &\bf{54.5} &53.4 &\bf{53.1} &45.1 &49.6 \\
    WiC &51.4 &\bf{52.6} &52.2 &\bf{47.8} &46.9 &47.3 &48.9 &50.1 &\bf{50.6} \\
    Multirc (F1a) &35.2 &40.6 &\bf{50.4} &57.1 &\bf{57.2} &52.8 &\bf{57.2} &48.2 &51.1 \\
    WSC &65.3 &70.2 &\bf{70.8} &66.7 &\bf{71.2} &71.5 &66.7 &\bf{70.2} &69.5 \\
    ReCoRD &75.8 &\bf{76.3} &75.3 &75.8 &\bf{76.4} &75.4 &74.9 &\bf{75.0} &74.6 \\
    CB &48.2 &\bf{50.0} &49.3 &44.6 &\bf{44.8} &42.9 &42.3 &\bf{48.2} &48.2 \\
    ANLI R1 &33.3 &33.5 &\bf{33.6} &31.3 &\bf{33.0} &32.2 &30.5 &\bf{32.5} &31.5 \\
    ANLI R2 &32.8 &\bf{34.2} &32.9 &30.5 &30.6 &\bf{31.6} &32.5 &\bf{33.4} &33.2 \\
    ANLI R3 &33.3 &\bf{33.6} &33.0 &30.0 &\bf{31.2} &31.0 &32.8 &\bf{34.2} &34.0 \\
    \midrule
    & \multicolumn{3}{c}{\bf Zero-shot} & \multicolumn{3}{|c|}{\bf One-shot} &  \multicolumn{3}{c}{\bf Few-shot} \\
    \cmidrule(l{3pt}r{3pt}r{3pt}){2-4} \cmidrule(l{3pt}r{3pt}r{3pt}){5-7} \cmidrule(l{3pt}r{3pt}r{3pt}){8-10}
    \bf Task & 
    PaLM \scriptsize{8B} &
    \ours \scriptsize{8B} &
    \oursext \scriptsize{8B} &
    PaLM \scriptsize{8B} &
    \ours \scriptsize{8B} &
    \oursext \scriptsize{8B} &
    PaLM \scriptsize{8B} &
    \ours \scriptsize{8B} &
    \oursext \scriptsize{8B}
    \\
    \midrule
Lambada (EM) &58.0 &\bf{59.1} &58.7 &65.8 &\bf{66.5} &66.2 &66.1 &\bf{67.5} &67.2 \\
StoryCloze &75.0 &\bf{75.6} &75.4 &75.0 &\bf{75.7} &75.1 &75.8 &\bf{76.2} &76.1 \\
PIQA &77.0 &\bf{77.4} &77.0 &75.5 &\bf{76.5} &76.2 &77.1 &\bf{77.3} &77.1 \\
ARC-e &55.3 &57.1 &\bf{58.1} &60.1 &\bf{60.2} &60.0 &64.0 &64.4 &\bf{65.4} \\
ARC-c &33.8 &33.0 &\bf{33.9} &34.0 &35.0 &\textbf{35.6} &35.5 &\bf{36.5} &36.1 \\
OpenbookQA &48.2 &49.2 &\bf{50.3} &47.0 &\bf{48.4} &48.0 &49.0 &\bf{49.5} &49.2 \\
Winograd &78.5 &\bf{80.6} &80.3 &79.5 &\bf{81.7} &81.3 &79.5 &81.2 &\bf{82.5} \\
Winogrande &60.0 &\bf{61.9} &61.5 &60.5 &\bf{62.1} &61.5 &61.0 &\bf{62.3} &62.0 \\
BoolQ &52.0 &\bf{62.1} &62.0 &53.7 &\bf{59.6} &55.9 &49.0 &57.7 &\bf{58.8} \\
Copa &82.0 &\bf{84.0} &83.8 &80.0 &83.0 &\bf{84.0} &82.0 &\bf{85.0} &84.5 \\
RTE &53.4 &\bf{53.9} &50.6 &\bf{55.2} &47.3 &48.1 &\bf{53.1} &48.4 &48.0 \\
WiC &51.3 &51.1 &\bf{51.5} &\bf{51.5} &46.9 &48.9 &50.5 &49.5 &\bf{50.6} \\
Multirc (F1a) &40.4 &54.1 &\bf{57.5} &49.8 &\bf{56.5} &55.8 &42.5 &46.5 &\bf{47.3} \\
WSC &78.3 &\bf{79.1} &79.0 &79.0 &\bf{86.8} &85.6 &77.9 &\bf{87.9} &86.3 \\
ReCoRD &85.5 &85.0 &\bf{85.7} &85.0 &84.9 &85.1 &\bf{84.6} &83.9 &84.4 \\
CB &50.0 &48.2 &48.1 &42.9 &\bf{51.5} &51.0 &46.4 &\bf{50.0} &48.8 \\
ANLI R1 &32.9 &\bf{34.3} &33.8 &32.7 &\bf{33.5} &33.2 &31.1 &\bf{32.9} &32.0 \\
ANLI R2 &33.3 &\bf{34.1} &33.0 &30.6 &\bf{33.7} &31.9 &31.7 &\bf{33.8} &32.9 \\
ANLI R3 &33.0 &\bf{33.9} &33.0 &31.7 &\bf{33.8} &32.0 &32.9 &\bf{35.1} &34.6 \\
    \bottomrule
    \end{tabular}
}
\end{table*}

\subsection{
Few-shot Performance
}
We compare \ours with PaLM on few-shot and zero-shot performance in a wide range of NLP tasks, including {LAMBADA}~\citep{paperno2016lambada}, {StoryCloze}~\citep{mostafazadeh2016corpus},
{PIQA}~\citep{bisk2019piqa}, {ARC}~\citep{yadav2019quick} (ARC-e (easy) and ARC-c (challenge)) {OpenBookQA}~\citep{mihaylov2018suit},
{Winograd} tasks~\citep{kocijan2020review}, {WinoGrande}~\citep{sakaguchi2020winogrande},  {SuperGLUE}~\citep{sarlin2020superglue}, {Adversarial NLI (ANIL)}~\citep{nie2019adversarial}.

Table~\ref{tab:few_shot_all_tasks} includes the results for the \ours and the PaLM 1B and 8B models.
The results averaged over task categories are presented in~Figure~\ref{fig:group_highlight}.

\ours outperforms PaLM on 17 out of 19 tasks in the zero-shot setting, 15 out of 19 tasks in the one-shot setting, and 15 out of 19 tasks in the few-shot setting.
On the SuperGLUE~\citep{sarlin2020superglue} benchmark, our method significantly improves the 1B-model-size PaLM's zero-shot performance from 55.7 to 59.2 and improves the 8B-model-size PaLM's zero-shot performance from 61.6 to 64.0.
We further observe that \oursext performs similarly to if not better than \ours, \eg, \oursext performs better than on zero-shot (though not on few-shot). This suggests that since in \oursext the first half of input sequence is exactly the same as \ours, it at least retains \ours's zero-shot and few-shot performance. 
Consider that PaLM is well-tuned %
in many aspects, including the pre-training dataset, training strategy, and the number of tokens observed. The significantly better results of \ours shows that the training objective can also play a crucial role in the model performance.

\subsection{
Finetuning Performance
}
We conduct finetuning experiments on the SuperGLUE benchmark to compare PaLM and \ours. 
Following PaLM experimental settings, models are finetuned with $5 \times 10^{-5}$ learning rate using the Adafactor optimizer. 
To reduce computation time, we use batch size 512 instead of the original batch size 32 in PaLM. %
The models are finetuned for 20K steps.

Table~\ref{tab:superglue_finetune} reports the \textit{validation} results on finetuning on task-proportionate mixture of SuperGLUE tasks. 
On SuperGLUE, we compare with state-of-the-art models such as T5 11B~\citep{raffel2020exploring} and UL2~\citep{tay2022unifying}, as well as PaLM~\citep{chowdhery2022palm} and show that \ours obtains significantly better performance than PaLM.
All models are trained on C4 dataset, T5 11B and UL2 are trained on 1000B tokens, the rest of models are trained on 180B tokens.
It is worth noting that both top performing models on SuperGLUE are encoder-decoder models that are trained using the span-corruption objective. 
It has been shown that such an architecture generally outperforms autoregressive decoder-only models on classification task finetuning, when training cost is equalized~\citep{raffel2020exploring}. 
These results demonstrate that \ours can help bridge the gap.
\ours 1B outperforms PaLM 1B significantly on 4 out of 8 SuperGLUE tasks, and \ours 8B significantly outperforms PaLM 8B on all 8 SuperGLUE tasks, improving the score from 80.7 to 83.1.
We further observe that \oursext boosts performance significantly, outperforming PaLM 1B by 6.5 percent and outperforming PaLM 8B by 7.1 percent. This suggests that \oursext can learn better representations from a bidirectional context.  

\begin{table*}[h!]
\setlength{\tabcolsep}{6pt}
\centering
\footnotesize
\caption{Finetuning results on SuperGLUE dev set. We compare with T5-11B~\citep{raffel2020exploring}, UL2~\citep{tay2022unifying} and PaLM~\citep{chowdhery2022palm}. Scores reported are the peak validation scores per task following the setup of T5. 
All models are trained on the same 180B tokens except that UL2 20B and T5 11B are trained on 1000B tokens.
}
\vspace{0.7em}
\label{tab:superglue_finetune}
\begin{tabular}{llc||cccccccc|c}
\toprule
\bf Type & \bf Model & \bf Size & BoolQ & CB & CoPA & MultiRC & Record & RTE & WiC & WSC & \bf Avg   \\
\midrule
\midrule
\multirow{3}{4em}{{Masked language model}} & T5 & 11B      & 90.8 & 94.9 & 98.0 & 87.4 & 93.8 & 93.9 & 77.3 & 96.2 & 89.9 \\
& UL2 & 20B & 90.8 & 98.7& 99.0 & 88.4 & 93.7 & 92.1 & 77.3 & 98.1 &  \textbf{90.7}  \\
& T5 & 1.4B & 83.7	& 92.9 &	85.9 &	82.7 &	69.6 &	78 &	80.8 & 	80 &	81.7 \\
\midrule
\midrule
\multirow{6}{4em}{{Causal language model}}& PaLM & 1B &  75.0 &	78.6 & 58.2	& 62.9 &	63.3	& 60.0 &	66.4 &	71.2  & 67.0 \\
& \ours & 1B & 73.5	& 82.0 &	58.0 &	63.0 &	65.1 &	61.0 &	66.5 & 80.8 & 68.7 \\
& \oursext & 1B &  75.7	& 84.9 & 67.2 & 68.0 & 71.8 & 68.0 & 72.9	& 79.8 & \bf{73.5} \\[0.25em]
& PaLM & 8B &83.7 &94.6 &80.0 &81 &71.2 &80.0 &75.2 &80.1 & 80.7 \\
& \ours & 8B &84.8 &96.4 &81.0 &82.1 &73.7 &86.0 &76.2 &85.0 &
83.1 \\
& \oursext & 8B & 87.2 & 98.8 & 90.6 & 86.8	& 80.1 & 96.0 & 75.8 & 86.8  & \bf{87.8} \\
\bottomrule
\end{tabular}
\vspace{-0.5em}
\end{table*}

\begin{table*}[h]
\caption{Comparisons on SuperGLUE between different models. The metrics are SuperGLUE zero-shot. 
PaLM$^\star$ denotes the published results of PaLM. 
}
\vspace{0.8em}
\label{tab:model_scale}
\centering
\setlength{\tabcolsep}{6pt}
\footnotesize
\begin{tabular}{llc||rrrrrrrr|r}\toprule
\bf Data & \bf Model & \bf Size & BoolQ &CB &COPA &MultiRC &ReCORD &RTE &WiC &WSC & \bf Avg \\\midrule
\midrule
\multirow{4}{4em}{High quality private data} & GPT-3 & 175B &60.5 &46.4 &91 &72.9 &90.2 &63.5 &0 &65.4 &61.2 \\
& PaLM$^\star$ & 540B &88 &51.8 &93 &83.5 &92.9 &72.9 &59.1 &89.1 &78.8 \\
& GPT-3 & 13B &66.2 &19.6 &84 &71.4 &89 &62.8 &0 &64.4 &57.2 \\
& PaLM$^\star$ & 8B &68.3 &41.1 &86 &47.5 &87.8 &54.2 &47 &78.9 & \bf{63.9} \\
\midrule
\midrule
\multirow{3}{4em}{C4 1000B tokens} & ST-MoE & 269B &40.8 &41.1 &56 &30.3 &50 &52.7 &50 &57.5 &47.3 \\
& T5-XXL & 11B &44.3 &37.5 &70 &23 &85.8 &48.8 &50.9 &59.3 &52.5 \\
& UL2 & 20B &63.1 &41.1 &85 &36.2 &88.1 &60.7 &49.8 &79.9 & \bf{63} \\
\midrule
\midrule
\multirow{6}{4em}{C4 180B tokens} & PaLM & 1B &45.9 &48.2 &72 &35.2 &75.8 &50.9 &51.6 &65.3 &55.6 \\
& \ours & 1B &56 &50 &74 &40.6 &76.3 &53.8 &52.4 &70.2 &59.2 \\
& \oursext & 1B & 54.3	& 49.3	& 74 & 50.4	& 75.3 & 58.2 & 52.2 & 70.8  & \bf 60.6 \\[0.25em]
& PaLM & 8B &52 &50 &82 &40.4 & 85.5 & 53.4 & 51.3 &78.3 &61.6 \\
& \ours & 8B & 62.1	& 48.2	&  84.0	& 54.1	& 85.0	& 48.0	& 51.1	& 79.1	& \bf{64.0} \\
& \oursext & 8B & 62.5	& 47.5	& 82.0& 56.9&  85.3& 47.2& 	50.2& 	79.5& 63.9 \\
\bottomrule
\end{tabular}
\vspace{-0.3em}
\end{table*}

\subsection{
Ablation Study%
}

\p{\ours works best with random ratio.}
We evaluate the impact of mask ratio on \ours using SuperGLUE zero-shot benchmark.
Table~\ref{tab:mask_ratio_study} presents the results of \ours with different mask ratios.
Among them, sampling random ratio between $[0.0, 0.15]$ performs significantly better than other choices. 
Sampling mask ratios from $0.0$ to $0.1$ or $0.15$ perform generally better than using fixed mask ratio $0.1$ or $0.15$, indicating that fixed mask ratios could potentially introduce pre-training and inference gap, and sampling random mask ratio is a simple way to alleviate it.
\begin{table*}[h]
\centering
\footnotesize
\caption{Ablation of mask ratio on SuperGLUE. 
Comparisons on SuperGLUE zero-shot and one-shot benchmark between fixed mask ratio and random mask ratios using \ours. The model size is 1B. 
\ours [$x$, $y$] denotes mask ratio is randomly sampled between $x$ and $y$.
}
\label{tab:mask_ratio_study}
\vspace{0.8em}
\begin{tabular}{ll||cccccccc|cc}
\toprule
\bf Shots & \bf Model &BoolQ &CB &COPA &MultiRC &ReCORD &RTE &WiC &WSC & \bf Avg \\\midrule
\midrule
\multirow{6}{*}{Zero-shot} & PaLM &45.9 &48.2 &72.4 &35.2 &75.8 &50.9 &51.6 &65.3 &55.7 \\
&\ours [0.1, 0.1] &56.5 &51.6 &73.5 &32.9 &76.3 & \bf{55.6} &52 &67.1 &58.2 \\
&\ours [0.15, 0.15] &54 &48.2 &75.5 &22.6 &75.9 &52.7 &49.8 &66.1 &55.6 \\
&\ours [0, 0.1] & \bf{57.9} &51.8 &69.6 &33.3 & \bf{76.8} &48.4 &51.6 &67.7 &57.1 \\
&\ours [0.0, 0.15] &56 &50 & \bf{74.1} &40.6 &76.3 & 53.8 & \bf{52.4} & \bf{70.2} & \bf{59.2} \\
&\ours [0.0, 0.3] &52.5 & \bf{53.6} &69.4 & \bf{42.9} &75.4 &49.8 &48.4 &66.1 &57.3 \\
\midrule
\midrule
\multirow{6}{*}{One-shot} &PaLM &48.3 &44.6 &50.9 &75.8 &47.8 &72 &35.9 &66.7 &55.3 \\
&\ours [0.1, 0.1] &56.1 &32.1 &53.8 &76.3 &47.3 &72 &33.3 & 67.4 &54.8 \\
&\ours [0.15, 0.15] &48.3 &37.5 &52.4 &75.9 &\bf{49.1} &69 &20.9 &66.3 &52.4 \\
&\ours [0, 0.1] &\bf{56.5} &42.9 &53.1 &\bf{76.8} &47.8 &72 &29.1 &66.3 &55.6 \\
&\ours [0.0, 0.15] &52.6 & \bf{44.6} & \bf{54.5} &76.4 &46.9 &73 &43.2 &\bf{71.6} & \textbf{57.9} \\
&\ours [0.0, 0.3] &50.8 &32.1 &52 &75.4 &47.2 &\bf{74} & \bf{46.5} &66.3 &55.5 \\
\bottomrule
\end{tabular}
\end{table*}

\p{Using mask tokens instead of attention mask.}
Alternative to \ours, a natural way of preventing future tokens from attending to past tokens is replacing tokens with a special \texttt{[mask]} token.
Using mask tokens is widely adapted in masked language modeling~\citep{devlin2018bert, liu2019roberta}, and combining mask token with causal language modeling can be considered as a special case of UniLM~\cite{dong2019unified}. We perform an ablation study comparing \ours with mask token, and present the results in Table~\ref{tab:mask_vs_attention}.
Using mask tokens lead to performance degradation in zero- and few-shot experiments, and about the same results on finetuning experiments. We hypothesis that the performance drop is due to the train and inference gap caused by introducing the \texttt{[mask]} token, which can negatively impact zero- and few-shot performance because the model is not finetuned to remove such gap.
\begin{table*}[h!]
\centering
\footnotesize
\caption{Comparisons on SuperGLUE zero-shot, few-shot and finetuning benchmarks between using attention mask and using mask token. The model size is 1B. 
}
\label{tab:mask_vs_attention}
\vspace{0.8em}
\footnotesize
\begin{tabular}{l||ccc|c}
\toprule
\bf Masking Method & 0-shot avg & 1-shot avg & 5-shot avg & finetune avg
\\\midrule
using \texttt{mask} token & 57.4 & 57.0 & 55.4 & 68.5\\
using attention mask & 59.2 & 57.9 & 57.2 & 68.7\\
\bottomrule
\end{tabular}
\end{table*}

\begin{table*}[!t]
\footnotesize
\centering
\caption{Comparisons on SuperGLUE zero-shot benchmark between between Random Masking vs. Dropout. The model size is 1B. 
}
\vspace{0.5em}
\label{tab:dropout_study}
\begin{tabular}{l||cccccccc|c}
\toprule
\bf Model &BoolQ &CB &COPA &MultiRC &ReCORD &RTE &WiC &WSC &\bf Avg \\\midrule
PaLM &45.9 &48.2 &72.4 &35.2 &75.8 &50.9 &51.6 &65.3 &55.7 \\
PaLM + \scriptsize{Dropout} &53.5 &48.2 &64.4 &37.2 &75.7 &50.2 &50.2 &63.5 &55.4 \\
\midrule
\ours \scriptsize{[0.1, 0.1]} + \scriptsize{Dropout} & 44	& \bf{53.6} &	71 &	\bf{43.1} &	75.3 &	\bf{59.2}	& 49.8 & 65.4 & 57.7 \\
\ours \scriptsize{[0.1, 0.1]} & \bf{56.5} & 51.6 & \bf{73.5} & 32.9 & \bf{76.3} & 55.6 & \bf{52} & \bf{67.1} & \textbf{58.2} \\
\bottomrule
\end{tabular}
\end{table*}

\p{Comparison with dropout.}
\ours random masking can be seen as a special type of dropout~\citep{Srivastava2014DropoutAS} applied only on the input sequence layer wisely by using attention masking. 
We note that general dropout and \ours are complementary in that they can be combined together. %
To compare random masking vs. dropout, we compare three models in Table~\ref{tab:dropout_study}: (1) PaLM, (2) PaLM with dropout rate 0.1, (3) \ours with fixed random ratio 0.1, and (4) \ours with fixed random ratio 0.1 and dropout rate 0.1.
We see that using dropout during large language models pre-training is harmful, decreasing the score from 55.7 to 55.4, which aligns with findings from prior work~\citep{raffel2020exploring, chowdhery2022palm}.
In contrast, combining dropout with \ours together can improve PaLM, improving the score 55.7 to 57.7, indicating that \ours and dropout are complementary techniques and we leave further studies of this as an interesting future work. 
We can see that using only \ours performs slightly better than combining dropout and \ours together, showing the effectiveness of \ours on performing the next token prediction task with randomly selected past tokens masked out. 

\section{Related work}
Large transformer models have made tremendous successes in natural language modeling. 
A number works demonstrated that training causal autoregressive models using next token prediction loss can achieve substantial improvements~\citep{dai2015semi, xie2017data, Peters2018DeepCW, radford2018improving, howard2018universal}. 
Later works further explore its effectiveness by scaling up the models and show impressive progresses in few-shot learning~\citep{brown2020language, radford2019language, rae2021scaling, hoffmann2022training, zhang2022opt}.
Meanwhile, substantial progresses have been made in training bidirectional encoders on masked language modeling~\citep{devlin2018bert, liu2019roberta, lewis2019bart}.
While causal autoregressive and bidirectional models have largely been developed as separate strains of work serving a different purpose, there have also been some attempts to combine the best of both worlds.
XLNet~\citep{yang2019xlnet} proposes to train causal models on all permutations of the factorization order, enabling the model to use bidirectional context and improve finetuning performance. Because of the permuted sequence, it requires a complex attention mechanism to avoid information leak, which doubles the time cost of pre-training. In contrast, \ours does not add extra computation.
Similarly, CM3~\citep{aghajanyan2022cm3} masks some spans that are moved to the end of sequence to be predicted. Unlike CM3, our approach focuses on improving models without altering the input sequence or changing the training objective. 
More studies have been conducted along this direction, sometimes in a simpler form, such as splitting documents into three pieces at random and moving the middle piece to the end~\citep{donahue2020enabling, du2022glm, bavarian2022efficient}.
\ours is orthogonal to these work and can be easily integrated into such methods.
\citet{raffel2020exploring} explores various training objectives for large encoder-decoder models.
HYBUNI~\citep{artetxe2022role} explores various masking strategies and training objectives, and conducts a comprehensive comparison of them.
Similarly,~\citet{wang2022language} conducts a comprehensive study of various language models and evaluates them on zero-shot learning and multi-task finetuning.
UniLM~\citep{dong2019unified, bao2020unilmv2} and UL2~\citep{tay2022unifying} propose to combine different training objectives together by using different self-attention masks to control the access to context for each token.
Our random masking is related to scheduled sampling~\citep{bengio2015scheduled} which replaces %
tokens with model predicted tokens and demonstrates improved sampling results. The most related works to \ours are
\citet{xie2017data, dai2015semi, bowman2015generating}, which study randomly masked tokens in training recurrent neural networks. %
In relation to them, our work focuses on efficiently improving transformer model by introducing masked language modeling to causal language modeling. Our approach does not need complex implementations to alter input or output sequence. In addition, it does not add extra computation.

\section{Conclusion}
In this paper, we propose \ours, a novel pre-training paradigm using a causal transformer decoder. 
\ours is a combination of causal next-token-prediction and random masking to input sequence. 
Experimental results show that \ours signiﬁcantly outperforms the state-of-the-art causal transformer model on a wide range of zero- and few-shot as well as finetuning benchmarks, and our model is readily extendable to various tasks.
In addition, \oursext %
is shown to further improve finetuning performance with similar few-shot results, making it potentially useful for real world applications where representations usage are ubiquitous, \eg, text search using embeddings. 

As \ours improves performance of causal language models on few-shot and finetuning benchmarks, applying our approach to other language understanding tasks and multimodal tasks is a promising direction for future work.
As \oursext introduces bidirectional context to causal language models, it would be an interesting and practical direction to further improve the finetuning performance using techniques from masked language modeling literature.

\section*{Acknowledgment}
We gratefully acknowledge the support from our colleagues in Google Research, including  Yi Tay for informative discussions and suggestions, Aakanksha Chowdhery for valuable help with PaLM training, Dale Schuurmans for important support, and Hanjun Dai for providing feedback on the paper.
We also would like to thank the support from teams in infrastructure and resource management.

\bibliography{main}
\bibliographystyle{icml2023}

\newpage
\appendix
\onecolumn
\section{Appendix}
\label{sec:appendix}
\subsection{Implementation and training details}
Our implementation uses Flax~\citep{flax2020github}, JAX~\citep{jax2018github} and T5X~\citep{roberts2022t5x}
Our architecture is based on PaLM~\citep{chowdhery2022palm} which introduces some modifications to GPT-3~\citep{brown2020language} architecture to reduce compute cost.
The dataset C4 is provided by Tensorflow datasets.
We use SentencePiece~\citep{kudo2018sentence} as tokenizer.
For PaLM and \ours trained on C4 datasets, the sequence length is 1024 to reduce compute cost, although the official PaLM is trained with longer context length 2048.
Following the settings of PaLM, input examples are concatenated together and then split into sequences of exactly 1024 tokens, so that there are no padding tokens, but examples may be split in the middle. Input examples are diﬀerentiated from one another with a special \texttt{[eod]} token
For downstream tasks evaluation, including both fewshot and finetune benchmarks, we follow the dataset format and splits used in~\citet{brown2020language, chowdhery2022palm}. 
Our experiments are conducted using cloud TPU v4, it has a unified 32 GiB HBM memory space across the entire chip.
For 1B models, training \ours on 180B tokens takes 25 hours on TPU v4-256 and training \oursext takes 51 hours. 
The training of \ours 8B models on 180B tokens takes 100 hours on TPU v4-512 and \oursext takes 140 hours.

\subsection{Hyperparameters}
In this section we provide the training and evaluation hyperparameters of \ours. These configurations follow the training hyperparameters of PaLM~\cite{chowdhery2022palm}.

\begin{table}[h!]
    \centering
    \begin{tabular}{l | l}
        \toprule
        \textbf{Hyperparameter} & \textbf{Value} \\
        \midrule
        Dropout & 0.0 \\
        Optimizer & Adafactor \\
        Initial learning rate $lr$ & 0.01 \\
        Learning rate decay & \makecell[l]{$0.01$ for the first 10,000 steps, \\ which is then decayed at a rate of $1/\sqrt{k}$, \\ where $k$ is the step number} \\
        Weight decay & $lr^2$ \\
        Optimizer momentum & $\beta_1 = 0.9, \beta_2 = 1.0 - k^{-0.8}$ \\
        Global norm gradient clipping & 1.0 \\
        Batch size & 1024 \\
        Sequence length & 1024 \\
        \bottomrule
    \end{tabular}
    \caption{Hyperparameters for training PaLM and \ours}
    \label{tab:train_hps}
\end{table}

\begin{table}[h!]
    \centering
    \begin{tabular}{l | l}
        \toprule
        \textbf{Hyperparameter} & \textbf{Value} \\
        \midrule
        Dropout & 0.1 \\
        Optimizer & SGD momentum \\
        Momentum & 0.9 \\
        Batch size & 512 \\
        Sequence length & 1024 \\
        \bottomrule
    \end{tabular}
    \caption{Hyperparameters for finetuning PaLM and \ours}
    \label{tab:eval_hps}
\end{table}

\subsection{Evaluation Tasks Details}\label{sec:tasks_details}
We consider the following tasks and categorize them according to their focused evaluation properties:
\begin{itemize}[leftmargin=15pt]
  \setlength\itemsep{0.1em}
    \item \emph{Cloze and Completion tasks}: \textbf{LAMBADA}~\citep{paperno2016lambada} consists of word prediction tasks that test the understanding of narrative passages. \textbf{StoryCloze}~\citep{mostafazadeh2016corpus} evaluates story understanding and script understanding, by requiring a system to choose the correct ending to a four-sentence story.
    \item \emph{Commonsense Reasoning}: \textbf{PIQA}~\citep{bisk2019piqa} is a dataset designed for physical commonsense reasoning to investigate the physical knowledge of language models. \textbf{ARC}~\citep{yadav2019quick} is a multiple-choice question-answering dataset, containing questions from science exams from grades 3-9. There are two partitioned datasets ARC-e (easy) and ARC-c (challenge), where the latter partition contains the more difficult questions that require reasoning. \textbf{OpenBookQA}~\citep{mihaylov2018suit} is designed to test understanding of both the topic (\eg, salient facts) and the language it is expressed in. This dataset contains questions that require multi-step reasoning, commonsense knowledge, and rich text comprehension.
    \item \emph{Winograd-style tasks}: In the Winograd schema challenge, a \emph{schema} is a pair of sentences that differ in only one or two words and that contain an ambiguity that is resolved in opposite ways in the two sentences. \textbf{Winograd} tasks~\citep{kocijan2020review} require world knowledge and reasoning to be solved. \textbf{WinoGrande}~\citep{sakaguchi2020winogrande} is a large-scale dataset of 44k problems, and requires commonsense reasoning to choose the correct option for a given sentence.
    \item \emph{Natural Language Understanding (NLU)}: \textbf{SuperGLUE}~\citep{sarlin2020superglue} consists of 8 challenging NLU tasks, including word sense disambiguation, natural language inference, coreference resolution, and question-answering.
    \item \emph{Natural Language Inference} (NLI): \textbf{Adversarial NLI (ANIL)}~\citep{nie2019adversarial} is collected via an adversarial human-and-model-in-the-loop procedure and is selected to be difficult to state-of-the-art models.
\end{itemize}

\subsection{Full results}\label{sec:full_result}
Table~\ref{tab:few_shot_all_tasks_three_evaluations} includes the results for the \ours and the PaLM 1B and 8B models across three random evaluation seeds.
Following prior work, we only consider single checkpoint results from pre-trained language models.
The variance across different evaluation seeds is small on most tasks.

\begin{table}[ht!]
    \footnotesize
    \caption{Results across three random realizations. 
    We use the same setup as in \citet{brown2020language, chowdhery2022palm}, including the splits for each task. 
    }
    \label{tab:few_shot_all_tasks_three_evaluations}
    \vspace{0.5em}
    \setlength{\tabcolsep}{4pt}
    \centering
    \footnotesize
    \begin{adjustbox}{angle=270}
    \begin{tabular}{p{2.0cm}|cc|cc|cc|cc}
    \toprule
    & \multicolumn{4}{|c|}{One-shot} &  \multicolumn{4}{c}{Few-shot} \\
    \cmidrule(l{3pt}r{3pt}r{3pt}r{3pt}){2-5} \cmidrule(l{3pt}r{3pt}r{3pt}r{3pt}){6-9}
    Task & 
    \makecell[c]{PaLM \\ \scriptsize{1B}} &
    \makecell[c]{\ours \\ \scriptsize{1B}} &
    \makecell[c]{PaLM \\ \scriptsize{8B}} &
    \makecell[c]{\ours \\ \scriptsize{8B}} &
    \makecell[c]{PaLM \\ \scriptsize{1B}} &
    \makecell[c]{\ours \\ \scriptsize{1B}} &
    \makecell[c]{PaLM \\ \scriptsize{8B}} &
    \makecell[c]{\ours \\ \scriptsize{8B}}
    \\
    \midrule
    Lambada (EM) & \makecell[c]{48.9 \\ \scriptsize{48.6 / 48.9 / 49.3}} & \makecell[c]{\bf{49.5} \\ \scriptsize{49.4 / 49.4 / 49.6}} & \makecell[c]{65.8 \\ \scriptsize{65.8 / 65.8 / 65.9}} & \makecell[c]{\bf{66.5} \\ \scriptsize{66.6 / 66.5 / 66.5}} & \makecell[c]{48.2 \\ \scriptsize{48.2 / 48.2 / 48.2}} & \makecell[c]{\bf{49.7} \\ \scriptsize{49.6 / 49.7 / 49.7}} & \makecell[c]{66.1 \\ \scriptsize{66.1 / 66.0 / 66.1}} & \makecell[c]{\bf{67.5} \\ \scriptsize{67.5 / 67.5 / 67.4}} \\
    StoryCloze & \makecell[c]{\bf{67.3} \\ \scriptsize{67.0 / 67.6 / 67.4}} & \makecell[c]{66.9 \\ \scriptsize{66.7 / 66.9 / 67.1}} & \makecell[c]{75.0 \\ \scriptsize{75.1 / 75.0 / 74.8}} & \makecell[c]{\bf{75.7} \\ \scriptsize{75.5 / 75.9 / 75.8}} & \makecell[c]{65.9 \\ \scriptsize{65.8 / 65.9 / 65.9}} & \makecell[c]{\bf{66.7} \\ \scriptsize{66.7 / 66.5 / 66.6}} & \makecell[c]{75.8 \\ \scriptsize{75.6 / 75.9 / 75.8}} &  \makecell[c]{\bf{76.2} \\ \scriptsize{76.0 / 76.5 / 76.3}} \\[0.5em]
    PIQA & \makecell[c]{71.0 \\ \scriptsize{71.0 / 71.0 / 71.0}} & \makecell[c]{\bf{71.6} \\ \scriptsize{71.7 / 71.6 / 71.6}} & \makecell[c]{75.5 \\ \scriptsize{75.5 / 75.4 / 75.6}} & \makecell[c]{\bf{76.5} \\ \scriptsize{76.5 / 76.5 / 76.5}} & \makecell[c]{\bf{72.0} \\ \scriptsize{72.0 / 72.1 / 72.1}} & \makecell[c]{71.6 \\ \scriptsize{71.6 / 71.6 / 71.7}} & \makecell[c]{77.1 \\ \scriptsize{77.1 / 77.1 / 77.1}} & \makecell[c]{\bf{77.3} \\ \scriptsize{77.3 / 77.3 / 77.3}} \\
    ARC-e & \makecell[c]{\bf{48.0} \\ \scriptsize{48.0 / 48.0 / 48.1}} & \makecell[c]{45.9 \\ \scriptsize{45.9 / 45.9 / 45.9}} & \makecell[c]{60.1 \\ \scriptsize{60.2 / 60.1 / 60.1}} & \makecell[c]{\bf{60.2} \\ \scriptsize{60.2 / 60.3 / 60.3}} & \makecell[c]{\bf{50.2} \\ \scriptsize{50.2 / 50.2 / 50.2}} & \makecell[c]{48.2 \\ \scriptsize{48.2 / 48.1 / 48.2}} & \makecell[c]{64.0 \\ \scriptsize{64.0 / 64.0 / 64.0}} & \makecell[c]{\bf{64.4} \\ \scriptsize{64.0 / 64.8 / 64.5}} \\
    ARC-c & \makecell[c]{26.3 \\ \scriptsize{26.3 / 26.6 /26.0}} & \makecell[c]{\bf{27.2} \\ \scriptsize{27.1 / 27.4 / 27.3}} & \makecell[c]{34.0 \\ \scriptsize{34.0 / 34.0 / 34.1}} & \makecell[c]{\bf{35.0} \\ \scriptsize{35.1 / 35.0 / 35.0}} & \makecell[c]{26.5 \\ \scriptsize{26.1 / 26.9 / 26.6}} & \makecell[c]{\bf{28.1} \\ \scriptsize{28.1 / 28.1 / 28.1}} & \makecell[c]{35.5 \\ \scriptsize{35.5 / 35.5 / 35.5}} & \makecell[c]{\bf{36.5} \\ \scriptsize{36.5 / 37.1 / 35.9}} \\
    \makecell[l]{Openbook\\-QA} & \makecell[c]{\bf{45.0} \\ \scriptsize{45.0 / 45.0 / 45.0}} & \makecell[c]{43.2 \\ \scriptsize{43.2 / 43.2 / 43.2}} & \makecell[c]{47.0 \\ \scriptsize{47.0 / 47.0 / 47.1}} & \makecell[c]{\bf{48.4} \\ \scriptsize{48.3 / 48.5 / 45.5}} & \makecell[c]{42.6 \\ \scriptsize{42.6 / 42.6 / 42.7}} & \makecell[c]{\bf{43.6} \\ \scriptsize{43.5 / 43.6 / 43.6}} & \makecell[c]{49.0 \\ \scriptsize{49.0 / 49.0 / 49.0}} & \makecell[c]{\bf{49.5} \\ \scriptsize{49.0 / 49.1 / 50.5}} \\[0.5em]
    Winograd & \makecell[c]{67.0 \\ \scriptsize{67.1 / 67.0 / 67.0}} & \makecell[c]{\bf{67.4} \\ \scriptsize{67.5 / 67.4 / 67.4}} & \makecell[c]{79.5 \\ \scriptsize{79.5 / 79.5 / 79.5}} & \makecell[c]{\bf{81.7} \\ \scriptsize{81.7 / 81.8 / 81.7}} & \makecell[c]{64.8 \\ \scriptsize{65.0 / 64.6 / 64.8}} & \makecell[c]{\bf{70.0} \\ \scriptsize{70.0 / 70.0 / 70.0}} & \makecell[c]{79.5 \\ \scriptsize{79.5 / 79.5 / 79.5}} & \makecell[c]{\bf{81.2} \\ \scriptsize{81.0 / 81.4 / 82.2}} \\
    Winogrande & \makecell[c]{54.0 \\ \scriptsize{54.0 / 54.0 / 54.0}} & \makecell[c]{\bf{55.8} \\ \scriptsize{55.8 / 55.8 / 55.9}} & \makecell[c]{60.5 \\ \scriptsize{60.1 / 60.5 / 60.9}} & \makecell[c]{\bf{62.1} \\ \scriptsize{62.1 / 62.2 / 62.1}} & \makecell[c]{53.6 \\ \scriptsize{53.6 / 53.6 / 53.7}} & \makecell[c]{\bf{55.0} \\ \scriptsize{55.9 / 55.8 / 53.3}} & \makecell[c]{61.0 \\ \scriptsize{59.1 / 58.9 / 65.0}} & \makecell[c]{\bf{62.3} \\ \scriptsize{62.3 / 62.4 / 62.3}} \\[0.5em]
    BoolQ & \makecell[c]{48.3 \\ \scriptsize{48.0 / 48.3 / 48.7}} & \makecell[c]{\bf{52.6} \\ \scriptsize{52.8 / 52.6 / 52.4}} & \makecell[c]{53.7 \\ \scriptsize{53.5 / 53.7 / 53.9}} & \makecell[c]{\bf{59.6} \\ \scriptsize{59.0 / 60.0 / 59.8}} & \makecell[c]{\bf{48.1} \\ \scriptsize{48.1 / 48.2 / 48.1}} & \makecell[c]{46.8 \\ \scriptsize{46.8 / 46.8 / 46.9}} & \makecell[c]{49.0 \\ \scriptsize{49.0 / 49.0 / 49.1}} & \makecell[c]{\bf{57.7} \\ \scriptsize{56.7 / 57.9 / 58.5}} \\
    Copa & \makecell[c]{72.0 \\ \scriptsize{71.0 / 73.0 / 74.0}} & \makecell[c]{\bf{73.0} \\ \scriptsize{73.0 / 73.0 / 73.0}} & \makecell[c]{80.0 \\ \scriptsize{79.0 / 80.0 / 81.0}} & \makecell[c]{\bf{83.0} \\ \scriptsize{82.0 / 84.0 / 83.0}} & \makecell[c]{70.0 \\ \scriptsize{70.1 / 71.0 / 70.0}} & \makecell[c]{\bf{72.0} \\ \scriptsize{72.0 / 71.0 / 73.0}} & \makecell[c]{82.0 \\ \scriptsize{83.0 / 81.0 /82.0}} & \makecell[c]{\bf{85.0} \\ \scriptsize{85.0 / 85.0 / 85.0}} \\
    RTE & \makecell[c]{53.1 \\ \scriptsize{52.0 / 53.1 / 54.2}} & \makecell[c]{\bf{54.5} \\ \scriptsize{54.5 / 54.5 / 54.5}} & \makecell[c]{\bf{55.2} \\ \scriptsize{55.0 / 55.2 / 55.4}} & \makecell[c]{47.3 \\ \scriptsize{47.3 / 47.3 / 47.3}} & \makecell[c]{\bf{53.1} \\ \scriptsize{53.0 / 53.1 / 53.1}} & \makecell[c]{45.1 \\ \scriptsize{45.1 / 45.1 / 45.1}} & \makecell[c]{\bf{53.1} \\ \scriptsize{53.1 / 53.1 / 53.1}} & \makecell[c]{48.4 \\ \scriptsize{48.4 / 48.4 / 48.3}} \\
    WiC & \makecell[c]{\bf{47.8} \\ \scriptsize{47.8 / 47.8 / 47.8}} & \makecell[c]{46.9 \\ \scriptsize{46.9 / 46.9 / 46.9}} & \makecell[c]{79.0 \\ \scriptsize{79.0 / 79.0 / 79.0}} & \makecell[c]{\bf{86.8} \\ \scriptsize{86.0 / 87.0 / 87.4}} & \makecell[c]{48.9 \\ \scriptsize{48.9 / 48.9 / 48.9}} & \makecell[c]{\bf{50.1} \\ \scriptsize{50.0 / 50.0 / 50.1}} & \makecell[c]{77.9 \\ \scriptsize{77.9 / 77.9 / 77.8}} & \makecell[c]{\bf{87.9} \\ \scriptsize{88.4 / 87.4 / 87.9}} \\
    Multirc (F1a) & \makecell[c]{57.1 \\ \scriptsize{57.1 / 57.1 / 57.0}} & \makecell[c]{\bf{57.2} \\ \scriptsize{57.0 / 57.4 / 57.2}} & \makecell[c]{49.8 \\ \scriptsize{49.8 / 49.8 / 49.9}} & \makecell[c]{\bf{56.5} \\ \scriptsize{56.9 / 56.0 / 56.6}} & \makecell[c]{\bf{57.2} \\ \scriptsize{57.2 / 57.3 / 57.1}} & \makecell[c]{48.2 \\ \scriptsize{48.2 / 47.9 / 48.5}} & \makecell[c]{42.5 \\ \scriptsize{42.3 / 42.3 / 42.9}} & \makecell[c]{\bf{46.5} \\ \scriptsize{46.5 / 46.2 / 46.8}} \\
    WSC & \makecell[c]{66.7 \\ \scriptsize{66.5 / 66.5 / 67.1}} & \makecell[c]{\bf{71.2} \\ \scriptsize{71.5 / 71.2 / 69.9}} & \makecell[c]{79.0 \\ \scriptsize{79.0 / 79.0 / 79.0}} & \makecell[c]{\bf{86.8} \\ \scriptsize{86.8 / 86.9 / 86.7}} & \makecell[c]{66.7 \\ \scriptsize{66.7 / 66.7 / 66.7}} & \makecell[c]{\bf{70.2} \\ \scriptsize{70.0 / 70.5 / 70.1}} & \makecell[c]{77.9 \\ \scriptsize{77.9 / 77.9 / 77.9}} & \makecell[c]{\bf{87.9} \\ \scriptsize{88.1 / 87.8 / 87.8}} \\
    ReCoRD & \makecell[c]{75.8 \\ \scriptsize{75.8 / 75.7 / 75.9}} & \makecell[c]{\bf{76.4} \\ \scriptsize{76.4 / 76.4 / 76.4}} & \makecell[c]{\bf{85.5} \\ \scriptsize{85.8 / 85.2 / 85.5}} & \makecell[c]{84.9 \\ \scriptsize{84.9 / 83.9 / 85.9}} & \makecell[c]{74.9 \\ \scriptsize{74.9 / 74.9 / 74.9}} & \makecell[c]{\bf{75.0} \\ \scriptsize{75.0 / 75.0 / 75.1}} & \makecell[c]{\bf{84.6} \\ \scriptsize{84.8 / 84.8 / 84.2}} & \makecell[c]{83.9 \\ \scriptsize{84.0 / 84.1 / 83.6}} \\
    CB & \makecell[c]{44.6 \\ \scriptsize{44.9 / 44.2 / 44.7}} & \bf{44.8} & \makecell[c]{42.9 \\ \scriptsize{42.9 / 42.9 / 42.9}} & \makecell[c]{\bf{51.5} \\ \scriptsize{50.5 / 52.0 / 52.0}} & \makecell[c]{42.3 \\ \scriptsize{42.3 / 42.3 / 42.3}} & \makecell[c]{\bf{48.2} \\ \scriptsize{48.1 / 48.1 / 48.4}} & \makecell[c]{46.4 \\ \scriptsize{46.6 / 46.6 / 46.0}} & \makecell[c]{\bf{50.0} \\ \scriptsize{51.0 / 51.0 / 48.0}} \\[0.5em]
    ANLI R1 & \makecell[c]{31.3 \\ \scriptsize{31.3 / 31.3 / 31.3}} & \makecell[c]{\bf{33.0} \\ \scriptsize{33.9 / 33.1 / 32.0}} & \makecell[c]{32.7 \\ \scriptsize{32.7 / 32.9 / 32.5}} & \makecell[c]{\bf{33.5} \\ \scriptsize{33.5 / 33.5 / 33.6}} & \makecell[c]{30.5 \\ \scriptsize{30.8 / 30.8 / 29.9}} & \makecell[c]{\bf{32.5} \\ \scriptsize{32.5 / 32.5 / 32.5}} & \makecell[c]{31.1 \\ \scriptsize{31.1 / 31.2 / 31.0}} & \makecell[c]{\bf{32.9} \\ \scriptsize{32.1 / 32.1 / 32.5}} \\
    ANLI R2 & \makecell[c]{30.5 \\ \scriptsize{30.6 / 30.6 / 30.3}} & \makecell[c]{\bf{30.6} \\ \scriptsize{30.4 / 30.4 / 31.0 }} & \makecell[c]{30.6 \\ \scriptsize{30.6 / 30.6 / 30.7}} & \makecell[c]{\bf{33.7} \\ \scriptsize{34.0 / 34.0 / 33.1}} & \makecell[c]{32.5 \\ \scriptsize{32.5 / 32.5 / 32.5}} & \makecell[c]{\bf{33.4} \\ \scriptsize{33.0 / 33.2 / 34.0}} & \makecell[c]{31.7 \\ \scriptsize{31.7 / 31.7 / 31.7}} & \makecell[c]{\bf{33.8} \\ \scriptsize{33.9 / 33.9 / 33.6}} \\
    ANLI R3 & \makecell[c]{30.0 \\ \scriptsize{30.2/ 30.2/29.6}} & \makecell[c]{\bf{31.2} \\ \scriptsize{31.0 / 31.1 / 32.5}} & \makecell[c]{31.7 \\ \scriptsize{31.5 / 31.5 / 32.1}} & \makecell[c]{\bf{33.8} \\ \scriptsize{33.5 / 33.5 / 34.4}} & \makecell[c]{32.8 \\ \scriptsize{32.6 / 32.6 / 33.2}} & \makecell[c]{\bf{34.2} \\ \scriptsize{34.0 / 34.0 / 34.6}} & \makecell[c]{32.9 \\ \scriptsize{32.8 / 32.7 / 33.2}} & \makecell[c]{\bf{35.1} \\ \scriptsize{34.1 / 35.9 / 35.3}} \\
    \bottomrule
    \end{tabular}
    \vspace{-0.5em}
    \end{adjustbox}
\end{table}

\newpage

\end{document}

%% file: math_commands.tex
\usepackage{amsmath,amsfonts,bm}

\def\eqref#1{equation~\ref{#1}}

\def\1{\bm{1}}

\DeclareMathAlphabet{\mathsfit}{\encodingdefault}{\sfdefault}{m}{sl}
\SetMathAlphabet{\mathsfit}{bold}{\encodingdefault}{\sfdefault}{bx}{n}